\documentclass[11pt,letterpaper]{article}
\usepackage[a4paper,
            bindingoffset=0.2in,
            left=1in,
            right=1in,
            top=1in,
            bottom=1in,
            footskip=.25in]{geometry}
\usepackage{tabularx,booktabs}
\usepackage{url}
\usepackage{fullpage}
\usepackage{graphicx}
\usepackage{lipsum}
\newcommand{\junk}[1]{}
\usepackage{amsmath,amssymb}
\usepackage[square,sort,comma,numbers]{natbib}
\usepackage{siunitx}
\usepackage{url}
\usepackage{algorithm}
\usepackage{algpseudocode}
\usepackage{amsmath}
\usepackage{amsfonts}
\usepackage[colorlinks,linkcolor=blue]{hyperref}

\usepackage{subfigure}
\usepackage{float}

\usepackage{multirow}
\usepackage{setspace}
\usepackage{mathrsfs}
\usepackage{authblk}
\usepackage{graphicx}
\usepackage{times}
\usepackage{subfiles}
\usepackage[left]{lineno} 
\usepackage{adjustbox}

\begin{document}
\thispagestyle{empty}

\title{Seeing Through Experts' Eyes: A Foundational Vision-Language Model Trained on Radiologists’ Gaze and Reasoning}


\author[1]{Kinhei Lee\thanks{These authors contributed equally to this work.}}
\author[1]{Peiyuan Jing$^*$}
\author[1]{Zhenxuan Zhang$^*$}
\author[1]{Yue Yang}
\author[1,2]{Tao Wang}
\author[3]{Dominic C Marshall}
\author[1]{Yingying Fang}
\author[1,4,5,6]{Guang Yang \thanks{Corresponding Author. Email: g.yang@imperial.ac.uk}}

\affil[1]{\small Bioengineering Department and Imperial-X, Imperial College London, London W12 7SL, UK}
\affil[2]{\small College of physics and information engineering, Fuzhou University, Fuzhou, 350108, China}
\affil[3]{Department of Surgery and Cancer, Imperial College London, London W12 0NN, UK}
\affil[4]{\small National Heart and Lung Institute, Imperial College London, London SW7 2AZ, UK}
\affil[5]{\small Cardiovascular Research Centre, Royal Brompton Hospital, London SW3 6NP, UK}
\affil[6]{\small School of Biomedical Engineering \& Imaging Sciences, King's College London, London WC2R 2LS, UK}

\date{}

\maketitle
\begin{abstract}
  Large-scale vision–language models have shown promise in automating chest X-ray interpretation. However, their clinical utility remains limited by a fundamental gap between model outputs and the reasoning processes of radiologists. Most systems optimize for semantic information without emulating how experts visually examine and interpret medical images. As a result, current models often overlook critical findings, misrepresent anatomical context, or generate reports that diverge from established diagnostic workflows. Radiologists typically follow structured protocols (e.g., the ABCDEF approach that sequentially assesses the airways, breathing, circulation, diaphragm, external and foreign material). This standardized workflow ensures that all clinically relevant regions are examined in a consistent and systematic manner. It reduces the risk of missed findings, supports reliable diagnostic reasoning and facilitates clear communication between radiologists and referring clinicians. Emulating such expert strategies is essential for improving the trustworthiness and interpretability of automated report generation systems. Therefore, we introduce Gaze-X, a vision–language model that leverages radiologists’ eye-tracking data as a behavioral prior to model expert diagnostic reasoning. By incorporating gaze trajectories and fixation patterns into pretraining, Gaze-X learns to follow the spatial and temporal structure of radiologist attention and integrates visual observations in a clinically meaningful sequence. This approach enables the model to align its focus with diagnostically relevant regions and to emulate the interpretive logic underlying expert reports. Using a curated dataset of gaze recordings, comprising over 30,000 key frames across diverse disease categories, from five radiologists interpreting chest X‑rays, we demonstrate that Gaze‑X produces more accurate, interpretable, and expert‑consistent outputs across a range of clinically relevant tasks and downstream applications. These include radiology report generation, disease grounding, and visual question answering, utilizing data from 231,835 radiographic studies with associated reports, 780,014 question–answer pairs, and 1,162 image–sentence pairs containing bounding boxes and corresponding phrases. Rather than solely improving output fluency or task performance, Gaze-X addresses a key barrier to clinical uptake by aligning model behavior with established radiological workflows. Specifically, unlike autonomous reporting systems, GazeX produces verifiable evidence artifacts, including inspection trajectories and finding-linked localized regions, enabling efficient human verification and safe human–AI collaboration. Learning “through expert eyes” thus provides a practical route toward more trustworthy, explainable, and diagnostically robust AI systems for radiology and beyond.
\end{abstract}


\newpage
\section{Introduction}

With the exponential growth in medical imaging volume and recent advances in multimodal generative artificial intelligence (AI), automated report generation from chest X-rays has become an increasingly important direction in clinical AI~\citep{tmiauto,naturefoundation,multilingual_report}. Such systems have the potential to reduce radiologists’ workload and improve the efficiency of medical reporting~\citep{tmiauto,yang2023radiology,gao2024simulating,naturefoundation}. While recent models demonstrate promising fluency in free-text generation, radiology reports are not merely descriptive narratives. They are structured diagnostic documents that reflect a step-by-step reasoning process grounded in anatomical and pathological interpretation. To ensure clinical reliability and interpretability, generated reports should follow established conventions in how radiologists observe, organize, and describe findings~\citep{structured_report,ai_in_rad}. This means that clinically useful and trustworthy report generation goes beyond image–text alignment. It must also follow standardized diagnostic protocols of clinicians~\citep{flamingocxr}.

\begin{figure}[!t]
    \centering
    \includegraphics[width=1\columnwidth]{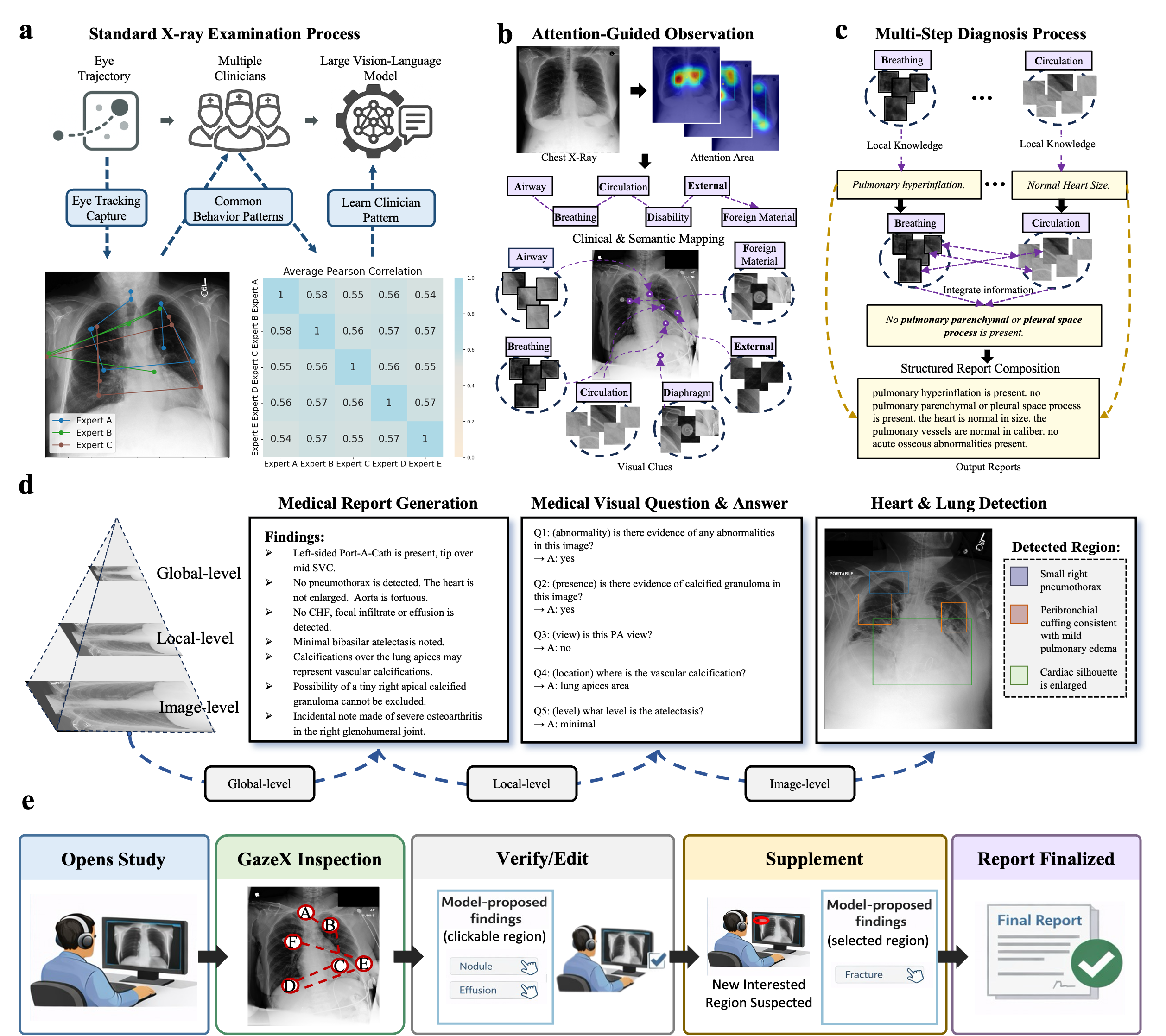}
    \caption{\textbf{Radiologist-guided diagnostic reasoning patterns for structured chest X-ray interpretation.} \textbf{a}, Radiologists’ gaze behaviors demonstrate structured and reproducible diagnostic workflows. Eye-tracking captures fixation trajectories from multiple experts, which reveal common attention patterns aligned with established reading protocols of clinicians. Statistical analysis of inter-expert trajectories shows consistent correlations, highlighting shared visual strategies during chest X-ray interpretation. Enabling large vision–language models to learn such standardized reading workflows has the potential to enhance the performance across diverse clinical downstream tasks. \textbf{b}, Attention-guided visual observations align with clinically defined categories (i.e., airways, breathing, circulation, diaphragm, external devices, and foreign materials), thereby providing a semantic mapping between image features (visual clues from different regions within the examined area) and diagnostic reasoning. \textbf{c}, Visual clues from key anatomical regions are integrated during multi-step diagnostic processes to support structured report generation. \textbf{d}, Downstream tasks such as segmentation, visual question answering (VQA), and report generation benefit from the model’s alignment with expert gaze. This figure illustrates how expert eye movement data capture the temporal and spatial logic of radiological decision-making, serving as supervision signals to guide AI models toward clinically grounded and interpretable report generation. \textbf{e}, Clinician-in-the-loop workflow enabled by GazeX. GazeX provides inspection cues and finding-linked regions to support systematic review and human verification, allowing radiologists to verify or edit model suggestions before finalizing the report.}
    \label{intro_eye}
\end{figure}

As shown in Fig.~\ref{intro_eye}, standardized guidelines for interpreting medical images are essential in clinical practice, as they ensure consistency across clinicians, support effective communication, and reduce the risk of overlooking critical findings~\citep{eye_review,modeling_eye}. In real practice, radiologists employ a systematic approach when interpreting chest X-ray images, often following the classic ABCDEF guideline (i.e., Airways, Breathing, Circulation, Diaphragm, External and Foreign material) to (1) minimize missed findings, (2) ensure diagnostic consistency, and (3) identify the patient’s most life-threatening problems first ~\citep{ChestXraysformedicalstudents,kool2007advanced}. Within this framework, specific observation patterns are followed for each area. For example, the lungs are divided into three zones (upper, middle, and lower), where each occupies about one-third of the lung’s height. Each zone is inspected from top to bottom for lung markings, which appear as small, faint white lines. Inspecting each zone in sequence helps radiologists systematically check for irregularities, such as missing markings, which may indicate a pneumothorax. More importantly, the sequential viewing approach, progressing from the entire lung to smaller subsections, allows radiologists to first assess whether the lung is uniformly expanded. When abnormalities are detected, radiologists can examine smaller anatomical regions to determine the underlying cause and confirm how extensive the pathology is. These examinations provide essential visual clues to verify and consolidate the diagnosis ~\citep{ChestXraysformedicalstudents}.
These viewing patterns reflect how radiologists allocate visual attention during image interpretation ~\citep{eye_mrl}. Recent studies have demonstrated that eye gaze trajectories are not random but closely follow clinical reasoning frameworks ~\citep{ai_human,flamingocxr}. However, current AI models lack this attention alignment. Incorporating radiologist eye gaze into training may help AI systems better emulate human diagnostic workflows and generate more clinically aligned reports.

Although many automated chest X-ray interpretation systems have emerged in recent years, they still face critical challenges in achieving the level of clinical reliability and consistency expected in real-world settings ~\citep{flamingocxr, dong2025keyword, gao2024simulating, wang2025survey,milam2023current,yu2023evaluating}. These challenges largely stem from the mismatch between how radiologists interpret medical images and how current AI models are designed ~\citep{dong2025keyword, li2024organ,hong2025diagnostic}. One major limit is the lack of clear alignment between the model’s diagnostic outputs and the corresponding anatomical regions in the image (e.g., when asked to identify pneumothorax, a system might return a positive response without indicating any visual evidence). The system’s conclusions remain opaque and difficult to verify without grounding its answers in observable image features \citep{r2gen, chen-etal-2021-cross-modal, yang2022media}. Another concern is whether the system thoroughly examines all relevant areas that a radiologist would routinely inspect. Radiologists are trained to follow a structured workflow, which includes checking the lung zones, heart borders, diaphragm contours, costophrenic angles, and soft tissues. Many AI systems do not have mechanisms to ensure such comprehensive coverage, often omitting regions like the lower lung fields where early signs of pathology (e.g., pleural effusions or basal infiltrates). This kind of partial examination can easily result in missed or inaccurate findings. In addition, current systems often lack the ability to integrate information across multiple regions of the image. Radiological diagnosis commonly requires synthesizing information from several anatomical areas \citep{CERROLAZA201944}. For instance, differentiating pulmonary edema from pneumonia often involves evaluating the symmetry of lung opacities, measuring the size of the cardiac silhouette, and assessing vascular congestion patterns. Without modeling such contextual dependencies, automated systems struggle to make accurate and coherent diagnostic judgments. These issues collectively raise concerns about the clinical applicability of end-to-end AI systems that operate without constraints. Therefore, it is essential to incorporate the standardized examination patterns and diagnostic reasoning processes that radiologists follow in practice into the design and training of automated systems to ensure safe and trustworthy deployment.

In this paper, we propose the GazeX framework. It leverages radiologists’ eye-tracking data to model the standard spatial and temporal dynamics of expert image interpretation. This helps bridge the gap between current AI models and expert radiologist behavior. The core idea of GazeX is to treat eye movement patterns as primary supervision signals that reflect clinically grounded reasoning strategies. During pretraining, the model is guided to attend to image regions in the order and manner that radiologists typically follow. This allows the system to learn not only which areas are diagnostically important but also how observations are integrated over time to support a final diagnostic decision. Compared to prior methods that focus on aligning generated text with report content or rely on region-level attention maps derived from class activation, GazeX is explicitly trained to reproduce expert gaze behavior. Our model internalizes detailed visual attention trajectories collected during real diagnostic workflows. This allows GazeX to move beyond the surface-level and rough alignment of free-form radiology reports with chest X-ray images and to approximate the deeper visual reasoning employed by radiologists. Moreover, while some prior studies have explored eye-tracking for prior information, our approach is the first to incorporate it directly into the model’s learning objective during pretraining. By embedding expert behavioral signals into the foundation of model training, GazeX offers a new paradigm for designing clinically grounded diagnostic systems. Rather than treating expert knowledge as a static label or external constraint, our method learns from how experts observe, integrate visual evidence, and reason through images during routine diagnosis. By aligning model behavior with established radiological workflows, this approach enables AI outputs to be reviewed and verified in relation to familiar reading steps and anatomical regions. This further facilitates clinician oversight and helps identify potential omissions. As a result, GazeX supports more transparent and interpretable human–AI interaction, providing a practical pathway toward trustworthy and expert-aligned AI systems for diagnostic radiology. As shown in Fig.~\ref{intro_eye}, we aim to model radiologists’ systematic visual inspection behavior to support clinician-in-the-loop chest X-ray interpretation.
(1) We analyzed radiologists' eye-tracking data to identify common patterns in how experts systematically inspect a chest X-ray image during diagnosis. This study involved five radiologists viewing the same chest X-ray images across 41 cases, with their fixations and trajectory correlations measured, revealing consistent inspection strategies across clinically relevant regions.
(2) We leveraged curated eye-tracking data comprising over 30,000 key frames spanning diverse disease categories to explicitly supervise model training in both temporal and spatial dimensions. This supervision enables the model to follow sequential attention patterns and to associate each reported finding with localized visual evidence, producing verifiable inspection trajectories and region-level evidence that can be audited by clinicians.
(3) We evaluated whether the model reads and interprets chest X-ray images in a manner consistent with expert radiologists, and whether enforcing systematic, inspection-trajectory constraints during model training improves coverage and reliability in downstream applications, as opposed to reliance on ungrounded shortcuts. These applications comprise a total of 231,835 radiographic studies with associated reports, 780,014 question–answer pairs, and 1,162 image–sentence pairs containing bounding boxes and corresponding phrases, spanning varying levels of complexity.

\section{Results}
\subsection{Evaluating GazeX performance on downstream tasks spanning different perspectives}
\begin{figure}[!htb]
    \centering
    \includegraphics[width=0.85\columnwidth]{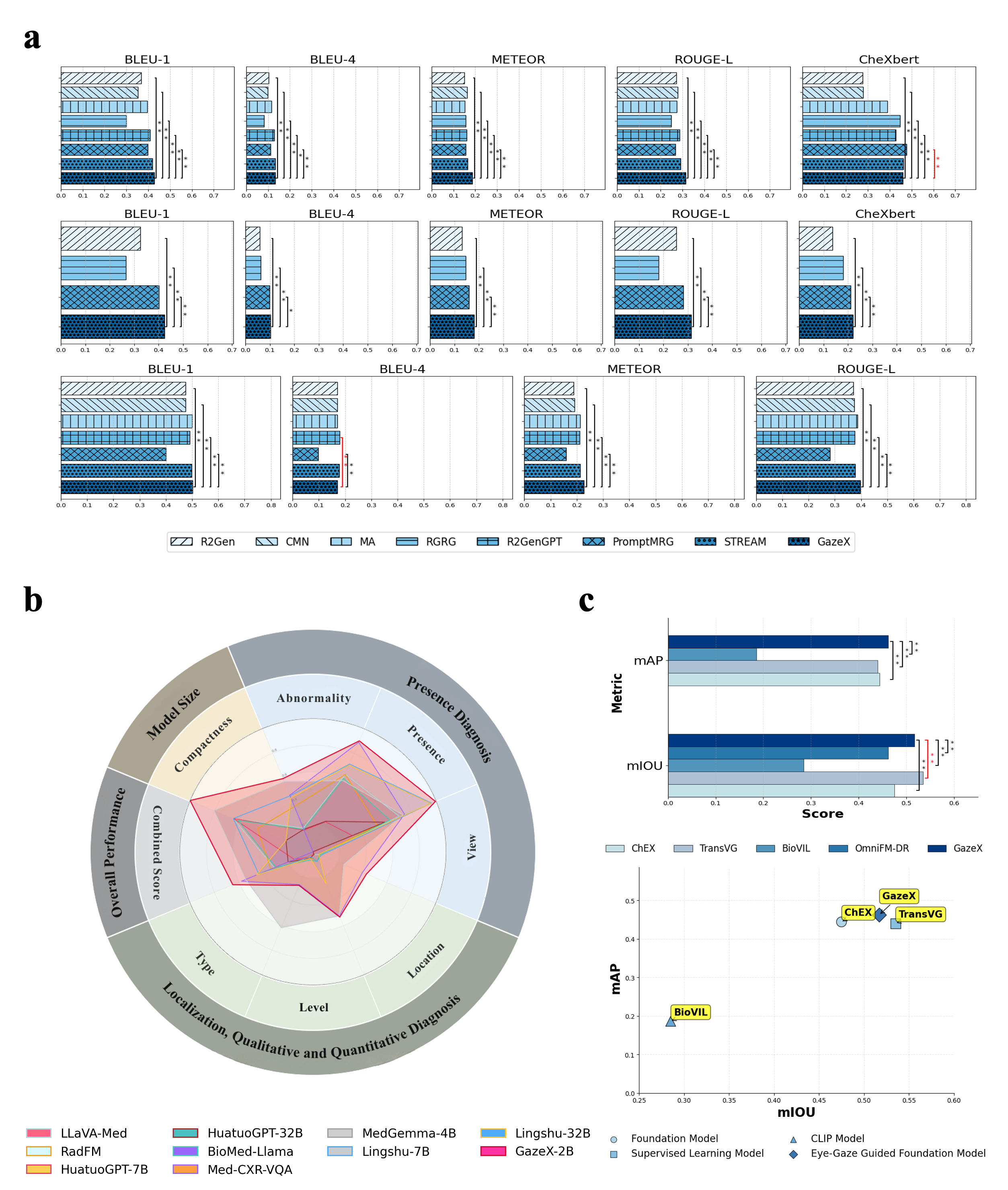}
    \caption{\textbf{Evaluating GazeX performance on chest radiology report generation, visual question answering and visual grounding.} \textbf{a}, Comparison of different report generation methods across various evaluation metrics in MIMIC-CXR~\citep{mimicxr} and IU X-ray~\citep{IUXray}. The shown results include evaluations on the MIMIC-CXR testing set using the GazeX model fine-tuned on MIMIC-CXR, zero-shot evaluations on the entire IU X-ray dataset using the GazeX model fine-tuned on MIMIC-CXR, and evaluations on the IU X-ray testing set using the model fine-tuned on the IU X-ray dataset. The Wilcoxon signed-rank test was performed to assess the statistical significance of the improvements compared to other models. \textbf{**} indicates $p < 0.01$, \textbf{*} indicates $p < 0.05$, and \textcolor{red}{\textbf{**}} indicates the reverse situation. \textbf{b}, Radar chart showing foundation models performance on different perspective of the visual question answering dataset, including model size, overall performance, localization, qualitative and quantitative diagnosis aspect and presence diagnosis aspect. \textbf{c}, Quantitative analysis, including mean Intersection over Union (mIOU) and mean Average Precision (mAP), of the accuracy of the visual grounding task across all disease categories for different model types (e.g., foundation models, CLIP-based models and generative models). Additional quantitative comparisons are provided in Supplementary Tables ~{\color{blue}1-3}.}
    \label{downstream_compare}
\end{figure}
\paragraph{GazeX performance in radiology report generation.} GazeX learns to emulate radiologists' global examination patterns, allowing it to integrate holistic visual information from the X-ray image for diagnosis and effectively transform these observations into free-form radiology reports. We conducted evaluation of GazeX report generation ability on two datasets: MIMIC-CXR (77,110 images and 227,835 reports) and IU X-ray (7,470 images and 3,955 reports). For the IU X-ray dataset, we evaluated two model versions: (1) GazeX model fine-tuned on the IU X-ray dataset and (2) zero-shot inference using the GazeX model pre-trained on MIMIC-CXR. The findings from \citep{rgrg} indicated that the official test split of the IU X-ray dataset is unsuitable for disease-specific evaluation due to insufficient positive samples for certain diseases. Given these limitations, for IU-Xray, clinical accuracy measured by CheXbert was assessed using the complete dataset instead of the designated test split for model version 2. As shown in Fig.~\ref{downstream_compare}a, GazeX demonstrates outstanding performance in NLG metrics, achieving state-of-the-art performance in BLEU and ROUGE-L scores across two report generation datasets: MIMIC-CXR ( 0.427 and 0.315 ) and IU X-ray setting (1) ( 0.502 and 0.398 ).  The model performance is stable across two public datasets with consistent improvements in BLEU-1 ( 0.427 in MIMIC-CXR and 0.502 in IU X-ray setting (1) ) and BLEU-2 ( 0.277 in MIMIC-CXR and 0.333 in IU X-ray setting (1) ) while maintaining BLEU-3 and BLEU-4 performance comparable to the state-of-the-art models, such as STREAM \cite{Stream}. Using the CheXbert-based F1 score as a factual accuracy metric, GazeX shows promising performance in generating clinically accurate radiology reports, achieving CheXbert F1 scores of 0.460 on MIMIC-CXR and the highest reported performance on IU X-ray (setting 2), highlighting its robust performance on the external dataset not used in training.
\paragraph{GazeX performance in visual-question answering.} GazeX not only learns to replicate the radiologist's global view analysis for a holistic diagnosis, but also adapts to targeted tasks by examining specific image regions for answering visual questions that require localized observations. The evaluation of visual-question answering is divided into six categories, Abnormality, Presence, Location, Level, Type, and View, each evaluating different levels of detail~\citep{medical_vqa_dataset}. For example, in chest radiograph analysis, multiple questions are posed when an abnormality is detected, including its type, severity, and anatomical localization. In the case of pulmonary edema, three questions arise: (1) the presence of edema, (2) its anatomical distribution, and (3) the degree of severity. Addressing these questions requires localized feature extraction and detailed observational analysis. As shown in Fig.~\ref{downstream_compare}b, GazeX demonstrates significant performance compared to established foundation models. We selected models pretrained on large-scale medical imaging datasets, including both exclusively radiological images and those with broader medical imaging data. While state-of-the-art foundation models like MedGemini, LLaVA-Med, and RadFM are pretrained on large-scale medical image-text datasets covering diverse tasks, GazeX utilizes radiologists' inspection patterns to achieve significant performance gains. When targeting specific diseases, GazeX can identify the anatomical regions that radiologists are most likely to examine. Furthermore, the model captures the comparative examination strategy employed by radiologists, who typically observe both abnormal areas and corresponding normal regions to establish diagnostic evidence and assess disease severity and presence. By learning from these visual inspection patterns, GazeX can more precisely identify abnormalities, severity levels and anatomical locations. GazeX exhibits significant improvements across multiple diagnostic categories quantitatively, including abnormality detection, presence identification, view classification, and anatomical localization. The model achieves the highest combined performance with a top-1 accuracy of 0.651, demonstrating the effectiveness of incorporating expert visual attention patterns in medical visual question answering tasks. In particular, when answering abnormality-related questions that require visual feature extraction from multiple regions and comparative analysis of normal and abnormal areas within chest radiographs, GazeX demonstrates state-of-the-art performance. Pre-trained in radiologist examination patterns, the model achieves a top-1 accuracy of 0.594 in abnormality identification and 0.902 in abnormality presence detection by effectively integrating both global and local features for accurate diagnostic reasoning. Global features encompass comparative analysis between the fields of the left and right lung to detect abnormalities such as edema, while local features focus on the detailed examination of specific abnormal regions within affected lung areas.

\paragraph{GazeX performance in visual grounding.} To perform a visual grounding task based on radiologists' findings, different types of methods are compared, including generative models ~\citep{Omnifm-dr}, CLIP-based models ~\citep{ms-cxr}, foundation models ~\cite{chex}, and supervised bounding box learning models ~\citep{transvg}. Our approach surpasses the foundation and CLIP-based models by an average of 42\% in mIOU and 60\% in mAP. GazeX replicates radiologist inspection patterns by learning associations between diseases and anatomical regions to identify critical visual indicators of abnormal findings. The system employs comparative analysis, contrasting normal and pathological regions (Fig.~\ref{discussion_comp}) or evaluating features across anatomical segments, such as lung lobes, to inform diagnostic decisions.
When identifying abnormalities in specific lung regions, GazeX compares the target area with normal regions for validation, mirroring standard radiological protocols. During fine-tuning for visual grounding tasks, GazeX generally adopts a broader observation area (Fig.~\ref{discussion_casestudy}f). Despite this, our model achieves performance comparable to supervised methods, which are explicitly tailored to model the relationship between radiologists' findings and annotated bounding boxes.

\section{Discussions}
\subsection{Radiologists exhibit consistent inspection patterns}
We leverage radiologist eye-tracking data collected during chest X-ray interpretation to pretrain GazeX with human-aligned visual supervision. To examine whether radiologists follow consistent inspection strategies, we analyzed 41 frontal chest X-ray images, each independently reviewed by five radiologists who provided clinical descriptions while their eye movements were tracked. Since raw gaze recordings may include fixations that fall outside the image boundaries, we further applied a filtering step that excluded these off-screen points.

\begin{figure}[!t]
    \centering
    \includegraphics[width=1\columnwidth]{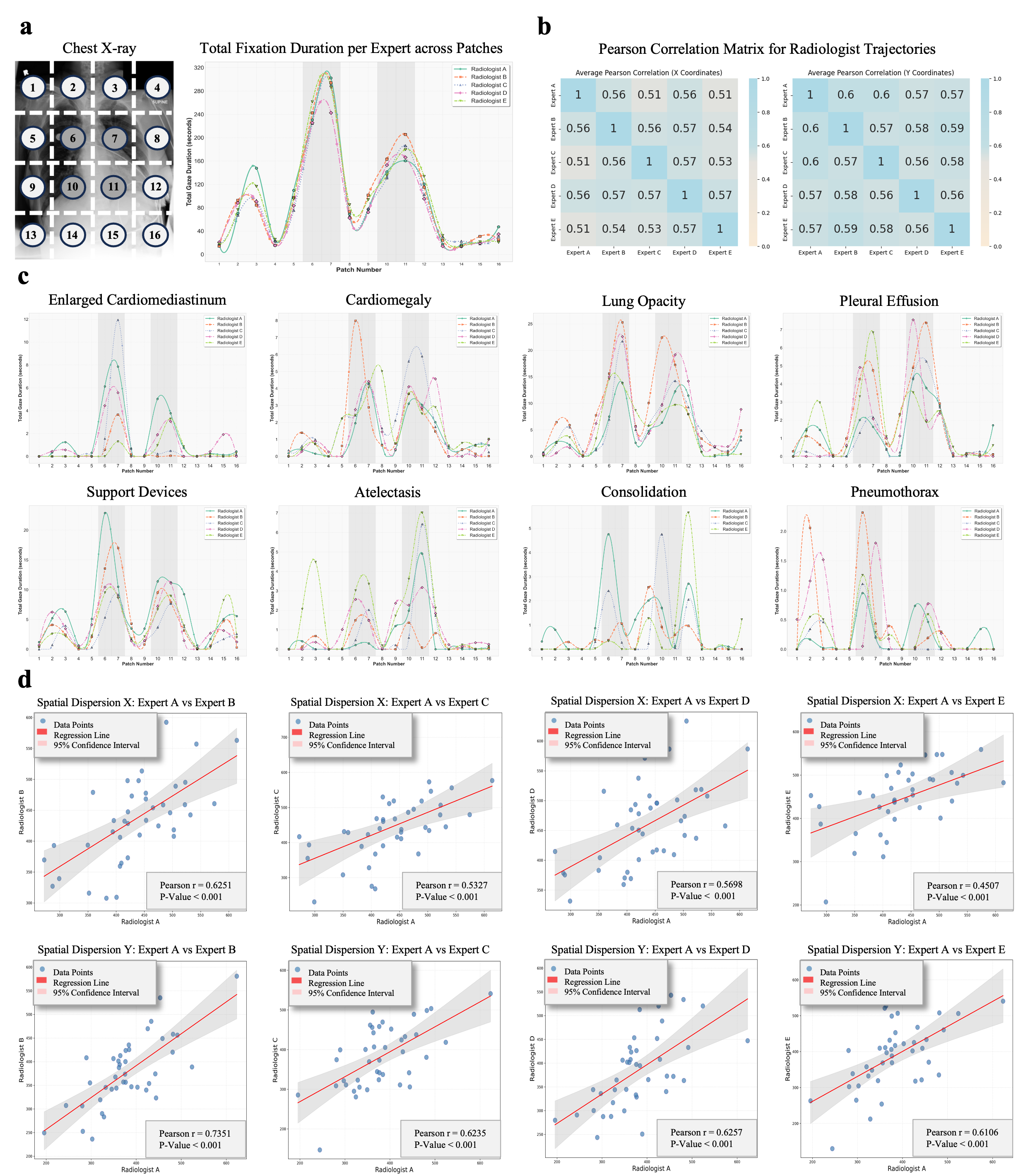}
    \caption{\textbf{Consistent inspection patterns among radiologists.} \textbf{a}, Total time spent per radiologist across a $4\times4$ grid of patches, illustrating consistent fixation patterns and dwell times, with the central thoracic regions receiving the greatest attention. \textbf{b}, Similarity of fixation trajectories quantified with Pearson correlation for the horizontal and vertical coordinates, indicating moderate spatial agreement. \textbf{c}, Total time spent per radiologist across a $4\times4$ grid of patches, stratified by representative case distributions across clinical categories. \textbf{d}, Case-level dispersion of fixation coordinates along both axes, revealing comparable spatial variability when attention is directed to similar anatomical regions.}
    \label{discussion_track}
\end{figure}

As shown in Fig.~\ref{discussion_track}, the processed gaze distributions reveal clear and repeatable attention patterns. Each image is divided into a 4×4 grid of equal-sized patches, and we calculate the cumulative gaze time within each patch between radiologists. Analysis for 6x6 and 8x8 grids are shown in Supplementary Fig.~{\color{blue} 1}. Patches in the central thoracic region, which correspond to the lungs and heart (patches 6, 7, 10, and 11) in 4x4 grid setting, consistently receive the longest dwell times, while peripheral areas are examined less (Fig.~\ref{discussion_track}a). Supplementary Fig.~{\color{blue}2} also shows that when comparing the viewing patterns of different radiologists across cases, there is a high correlation in how their views are dispersed within each case. This concentration of attention reflects established diagnostic practice. The lungs and heart are the primary areas where radiologists expect to identify clinically significant findings, including opacities, effusions, cardiomegaly, and vascular changes~\citep{cid2024development}. We also analyzed clinical categories encompassing both disease findings and support devices, with gazing time recorded using the same patch-based system shown in Fig.~\ref{discussion_track}c. For example, when assessing lung opacity, radiologists examined both lungs across the parenchyma, whereas for pneumothorax, attention was concentrated on the apical and peripheral regions where this pathology typically manifests. Correlation analyses of gaze trajectories across radiologists yield average Pearson coefficients of 0.55 for the horizontal axis and 0.58 for the vertical axis, indicating moderate spatial agreement (Fig.~\ref{discussion_track}b). Case-level dispersion plots further confirm shared temporal dynamics in the scanning behavior (Fig.~\ref{discussion_track}d). These findings show that radiologists use consistent patterns when viewing images, following routines from their clinical training. The data reveals valuable insights for developing AI models that can inspect X-rays like radiologists.

\begin{figure}[!t]
    \centering
    \includegraphics[width=1\columnwidth]{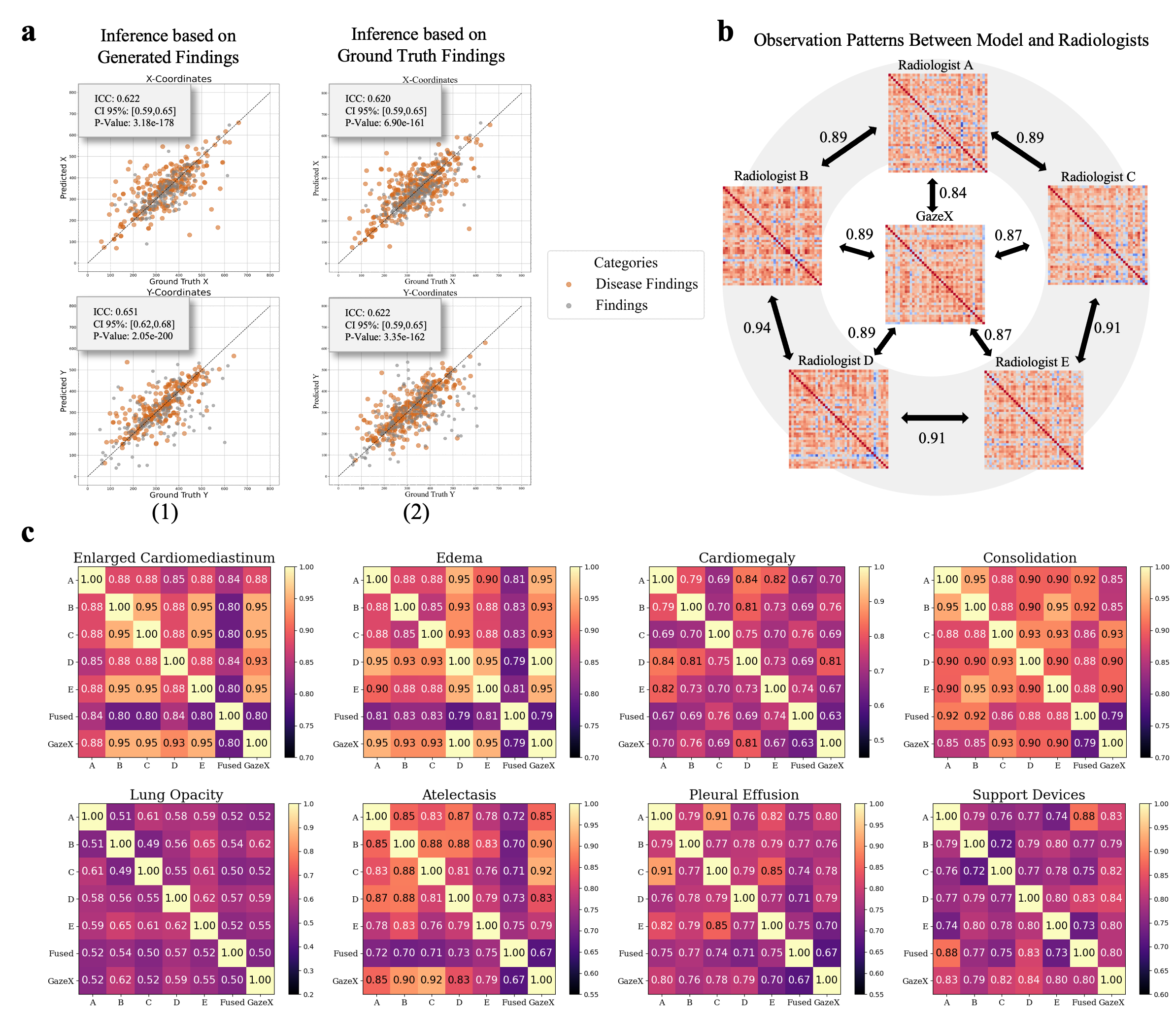}
    \caption{\textbf{GazeX emulates radiologists’ inspection patterns across spatial, temporal, and disease-specific contexts.} \textbf{a}, Scatter plots showing agreement between GazeX predicted gaze centroids and those from radiologists, quantified by intraclass correlation coefficients (ICC) for X and Y coordinates under two evaluation settings. \textbf{b}, Pairwise cosine similarity of attention maps derived from gaze clusters, comparing GazeX with each radiologist and inter-radiologist pairs. High similarity values indicate that the model replicates individual viewing behaviors with fidelity comparable to expert–expert agreement. \textbf{c}, Disease-specific similarity analysis across the 14 CheXbert categories. The fused category heatmap was computed by averaging the heatmaps from five radiologists. For most conditions, GazeX achieves correlation levels on par with, or approaching, the highest inter-radiologist similarity, and remains consistent with aggregated radiologist patterns.}
    \label{discussion_model}
\end{figure}

\subsection{GazeX reproduces radiologist inspection patterns}
After pre-training, GazeX can reproduce radiologists’ inspection trajectories, attend to salient gaze clusters, and generate findings informed by expert reading behavior. We evaluate this ability under two experimental settings (Fig.~\ref{discussion_model}a), using the same set of chest X-ray cases and gaze annotations analyzed in the previous subsection. For each disease category, a separate scatter plot is provided in Supplementary Figs.~{\color{blue}3-4}.

In the first setting (Fig.~\ref{discussion_model}a(1)), GazeX generates both the findings and their associated gaze centroids, simulating a realistic deployment scenario. CheXbert is applied to match generated and reference findings within the same diagnostic category, forming comparable pairs. Since the number of generated findings may differ from the reference findings, the corresponding gaze points may also vary in number. To address this, we used Dynamic Time Warping to align centroid sequences while preserving their temporal order for ICC analysis.

In the second setting (Fig.~\ref{discussion_model}a(2)), ground truth findings are provided as input to guide centroid extraction, ensuring that spatial comparisons are performed on identical findings. In both settings, GazeX and radiologists are treated as fixed raters, and their spatial agreement is quantified using ICC(3,1), a two-way mixed effects model with absolute agreement. The results indicate moderate spatial agreement under both conditions: for centroids derived from generated findings, ICC\_X = 0.622 and ICC\_Y = 0.651; for centroids derived from ground truth findings, ICC\_X = 0.620 and ICC\_Y = 0.622. These values suggest that GazeX achieves a spatial alignment level with radiologists that is comparable to agreement levels commonly reported in human–human reliability studies, reflecting a meaningful overlap in fixation patterns.

Inspired by the Representational Similarity Analysis described in \citep{du2025human,doerig2025high}, we evaluated the similarity of overall viewing patterns by transforming centroids and bounding boxes from both GazeX and radiologists into anisotropic Gaussian attention maps. Each Gaussian was centered at its corresponding centroid, with variance determined by the dimensions of the bounding box. Rather than generating heatmaps directly from the raw radiologists’ gaze data, we employed this approach to mitigate noise inherent in eye-tracking. The centroid and bounding box were derived using a weighted averaging procedure, as described in Methods 4.1, thereby capturing the region of highest visual concentration. For each case, we calculated the cosine similarity between the attention maps, generating heatmaps that capture individual inspection styles. As illustrated in Fig.~\ref{discussion_model}b, inter-radiologist inspection behaviors exhibit remarkable consistency, with average similarity scores exceeding 90\%. Notably, GazeX demonstrates robust behavioral emulation capabilities, achieving high similarity scores with radiologist B and radiologist D (0.89 and 0.89), who themselves exhibit the highest inter-radiologist similarity among all pairs (0.94). These behavioral patterns remain consistent when subjected to disease-based analysis, suggesting that the observed similarities are not confined to general inspection behaviors but extend to pathology-specific examination strategies.


Following the similar approaches, we conducted disease-specific analysis across the 14 CheXbert categories. From Fig.~\ref{discussion_model}c, we selected eight categories with comparatively representative sample sizes for analysis. The analysis further shows that, for most conditions, GazeX maintains correlation patterns with individual radiologists that are consistent with expert inspection strategies. Across disease categories, GazeX demonstrated consistent and stable performance compared to individual radiologists. For example, in the 'Edema' category, GazeX's inspection patterns closely resembled those of Radiologist D, with similarity ranges comparable to, or even exceeding, inter-radiologist variations. Notably, Radiologist D also exhibited high similarity to the other four radiologists, suggesting that GazeX may have captured commonly used inspection patterns among expert clinicians. However, in cases with varied inspection patterns, like 'Lung Opacity,' our model may not identify strong commonalities in observation patterns, yet still demonstrates generalizability across radiologists.

\begin{figure}[!t]
    \centering
    \includegraphics[width=1\columnwidth]{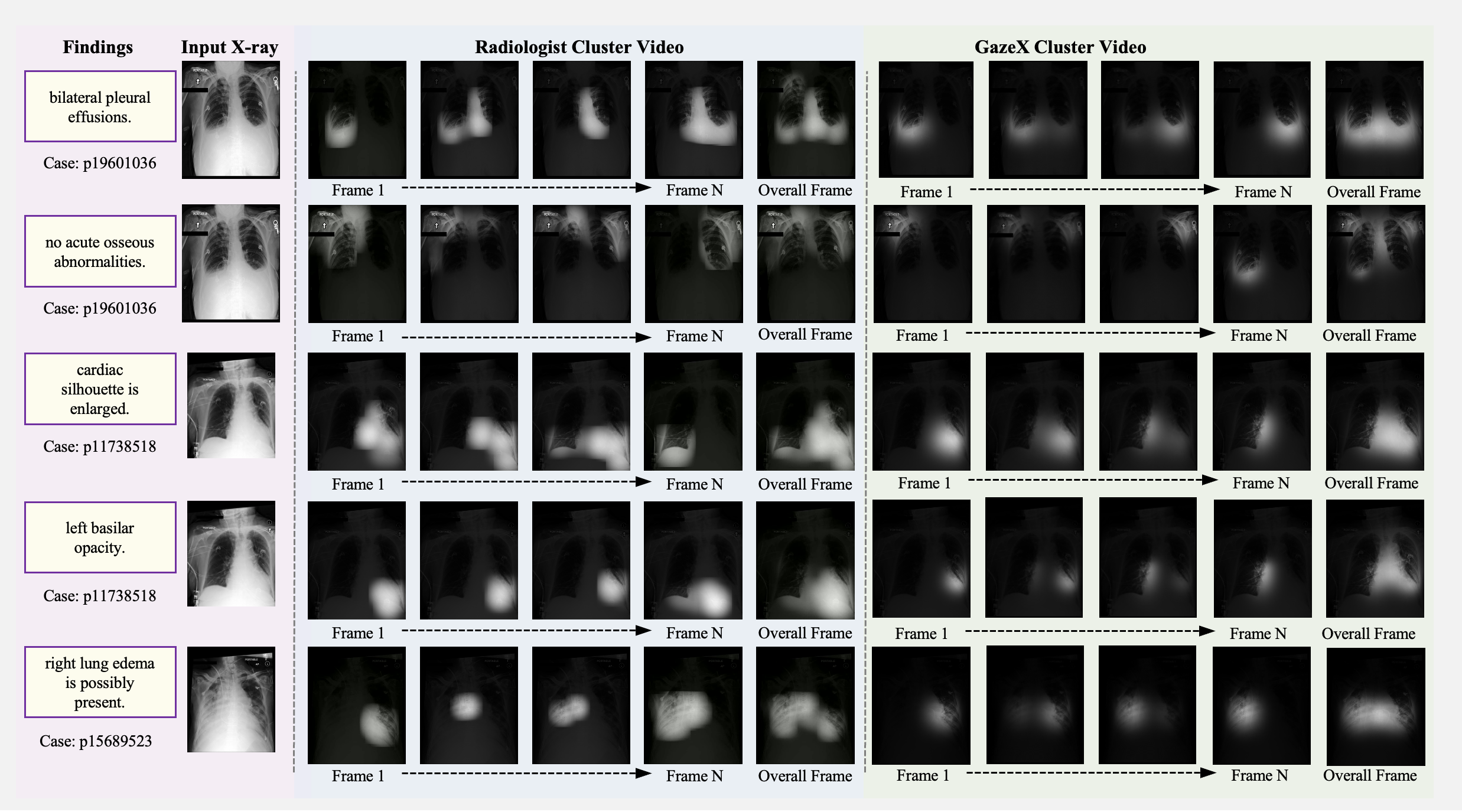}
    \caption{\textbf{Case-level comparison of disease-specific inspection patterns between GazeX and radiologists.} Representative examples from five CheXbert disease categories showing model-generated inspection sequences (right) alongside corresponding radiologists' inspection patterns (middle).}
    \label{discussion_comp}
\end{figure}

Representative examples from five disease categories are presented in Fig.~\ref{discussion_comp}. In each example, GazeX’s predicted the gaze clusters closely match the spatial distribution and temporal progression of radiologists' gaze patterns for the same clinical finding. For instance, when examining bilateral pleural effusions, both radiologists and GazeX concentrate their attention on the lung bases while comparing bilateral regions. In cases of 'left basilar opacity,' the model follows a similar pattern to radiologists by first focusing on the abnormal left basilar area, then examining the corresponding right-sided region to identify potential abnormalities, illustrating a systematic comparative methodology. These qualitative results complement the quantitative correlations, providing visual confirmation that GazeX can emulate radiologist-like inspection behavior in both spatial targeting and temporal sequencing. Representative disease-specific gaze-guided attention videos are shown in Supplementary Figs. ~{\color{blue}5–16}.

\subsection{Contribution of inspection pattern modeling}
To evaluate how each component contributes to the model's inspection capability, we tested GazeX with various ablated modules in both zero-shot and LoRA fine-tuned settings, assessing diagnostic accuracy (MIMIC-CXR reports) and attention alignment (REFLACX gaze data).
\begin{figure}[!t]
    \centering
    \includegraphics[width=1\columnwidth]{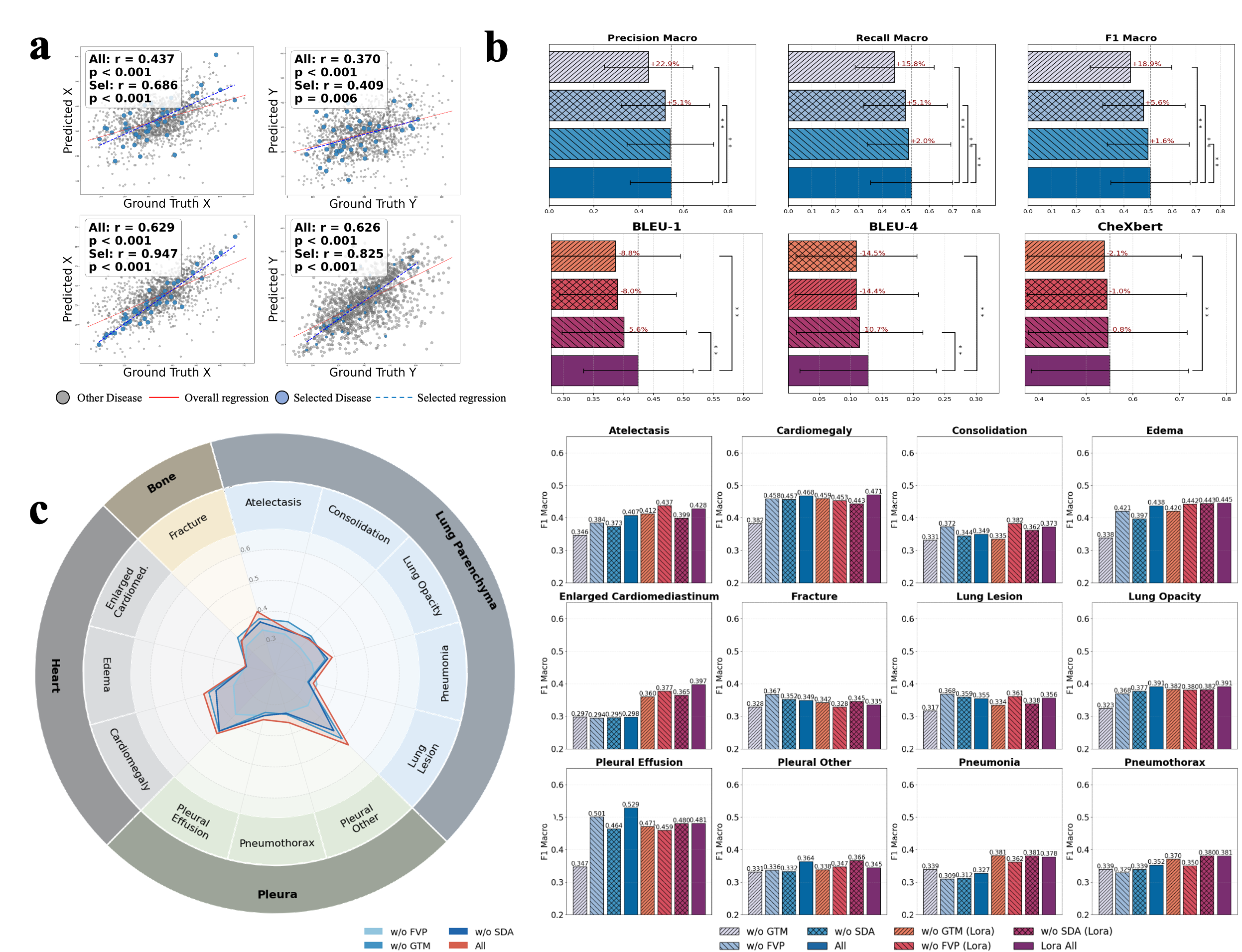}
    \caption{\textbf{Quantitative evaluation of GazeX against baseline and ablated variants.} \textbf{a}, Correlation between predicted and ground-truth gaze centroids for GazeX and the baseline Qwen2-VL without eye-tracking guided pretrain, reported as Pearson correlation coefficients (r) with corresponding p-values. Results are shown for all disease categories combined and for selected clinically significant categories. \textbf{b}, Ablation study results. The top row reports classification metrics (Precision, Recall, and F1) for GazeX without LoRA adaptation. The bottom row reports report-generation metrics (BLEU-1, BLEU-4) and clinical correctness (CheXbert F1) for fine-tuned GazeX with LoRA adaptation, under progressive removal of core components (FVP, GTM, and SDA). \textbf{c}, Condition-specific ablation results for 12 representative medical categories spanning lung, pleura, heart/mediastinum, bone/chest wall, and device-related findings. The radial plot summarizes the overall relative performance of GazeX before downstream task adaptation across disease categories, excluding Support Devices. Bars indicate the comparison between all ablated modules for GazeX and LoRA fine-tuned GazeX for the report generation task.}
    \label{discussion_ablation}
\end{figure}

We first compare GazeX with baseline, Qwen2-VL, to verify whether radiologist inspection patterns are effectively learned and exploited. In fairness, both models are provided with ground-truth findings to generate corresponding attention regions. Since Qwen2-VL lacks native sentence-level localization, we fine-tuned it using bounding boxes derived from radiologist gaze clusters, mirroring our Fine-Grained Visual Perception module. On the REFLACX test set (Fig.~\ref{discussion_ablation}a), GazeX achieved substantially higher correlation with radiologist gaze clusters ($r_X=0.629$, $r_Y=0.626$) than Qwen2-VL ($r_X=0.437$, $r_Y=0.370$). Gains are especially pronounced in clinically important categories such as support devices, lung lesion, enlarged cardiomediastinum, pleural other, fracture, and pneumothorax (GazeX: $r_X=0.947$, $r_Y=0.825$; Qwen2-VL: $r_X=0.686$, $r_Y=0.409$).

We then perform an ablation study to quantify the contribution of the Fine-Grained Visual Perception (FVP), Gaze Trajectory Mimicking (GTM), and Sequential Dependency Awareness (SDA) modules (Fig.~\ref{discussion_ablation}c). Removing any component reduces classification metrics (Precision, Recall, F1), and even without task-specific fine-tuning, the full GazeX model which driven only by learned inspection patterns, demonstrates strong diagnostic understanding and consistently outperformed all ablated variants. Incorporating LoRA adaptation \citep{lora} for these modules further boosts report generation metrics (BLEU-1, BLEU-4) and clinical correctness (CheXbert F1). The largest performance drops occur when GTM or SDA are removed, confirming their central role in gaze-informed reasoning. Wilcoxon signed-rank tests show that integrating all three modules yields statistically significant gains ($p<0.01$).

Condition-specific analysis demonstrates that GazeX maintains strong performance across the 14 CheXbert diagnostic categories (Fig.~\ref{discussion_ablation}d). The radial plot summarizes relative performance for 13 categories, excluding No Findings, while the accompanying bar charts report CheXbert F1 scores for 12 categories, excluding No Findings and Support Devices. Even without task-specific fine-tuning, GazeX matches or surpasses fine-tuned baselines in several clinically important conditions, with particularly notable gains in lung opacity and pleural effusion. In most categories, both the full GazeX model and its LoRA-adapted variant outperform the ablated configurations, indicating that inspection pattern modeling contributes not only to improved overall accuracy but also to consistent robustness across diverse disease types.

\begin{figure}[!t]
    \centering
    \includegraphics[width=1\columnwidth]{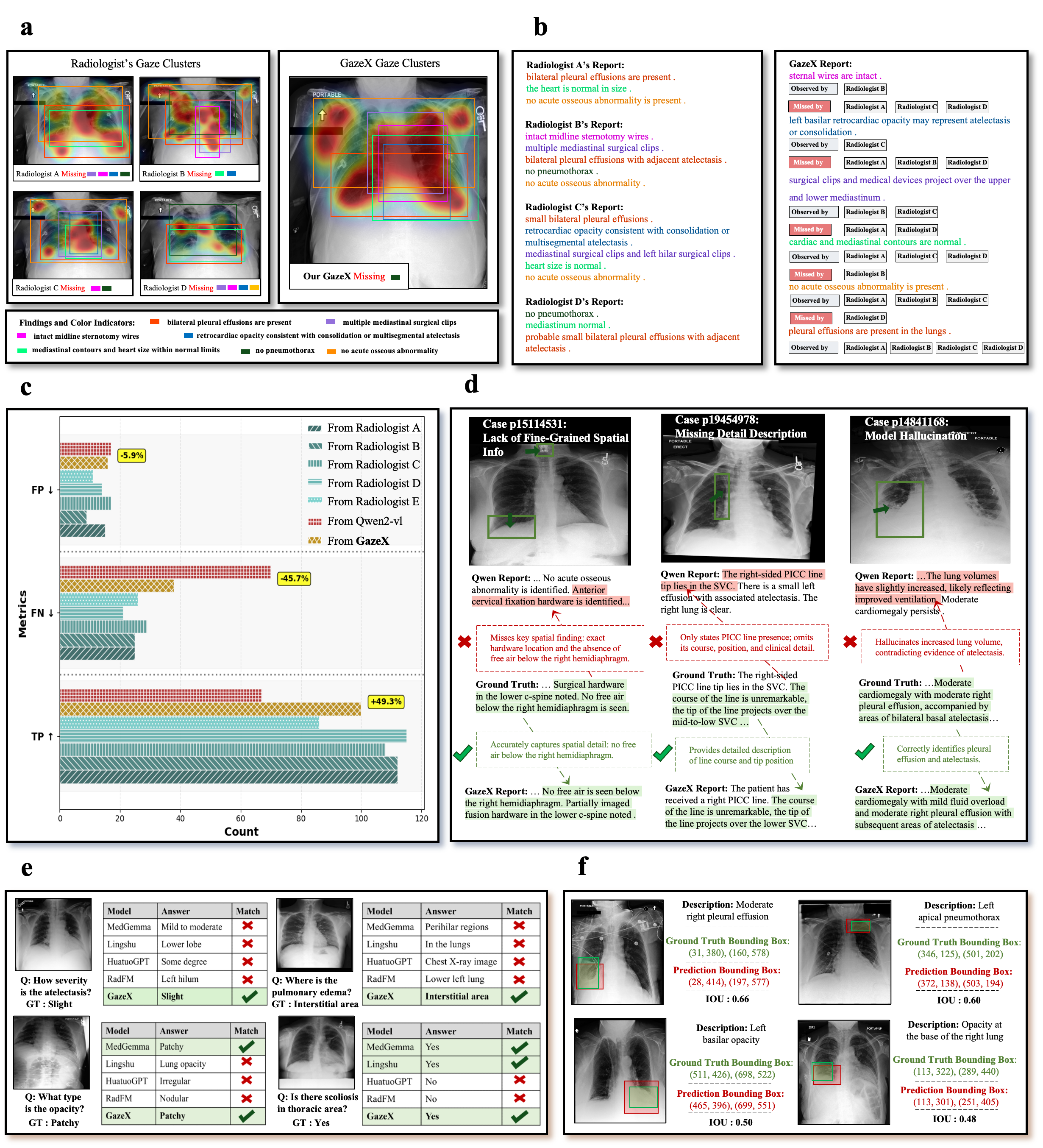}
    \caption{\textbf{Case study analysis demonstrating the advantages of inspection-guided modeling in GazeX.} \textbf{a}, Comparison of radiologists’ gaze clusters and GazeX-predicted gaze clusters on the same case. Colored bounding boxes indicate clinically relevant findings, with “Missing” labels marking unobserved findings. \textbf{b}, Alignment between GazeX-generated reports and individual radiologists’ reports. Colored highlights indicate matched findings; “Observed by” and “Missed by” tags quantify agreement with each radiologist. \textbf{c}, Quantitative comparison of false positives (FP), false negatives (FN), and true positives (TP) for GazeX, Qwen2-VL baseline, and individual radiologists, using majority-vote consensus reports as reference. \textbf{d}, Qualitative error analysis comparing baseline and GazeX reports. \textbf{e}, Visual question answering comparisons across multiple medical vision–language models. \textbf{f}, Localization performance for descriptive findings, with predicted bounding boxes and Intersection-over-Union (IoU) values against ground-truth annotations.}
    \label{discussion_casestudy}
\end{figure}

\subsection{GazeX augments radiologist findings through collective inspection patterns}

GazeX can potentially identify clinically relevant findings that individual radiologists may not mention. As illustrated in Fig.~\ref{discussion_casestudy}a, GazeX generates findings and highlights regions of interest based on the collective inspection patterns learned from all radiologists. This enables GazeX to detect abnormalities that some individual experts omit, while also aligning its attention with regions commonly examined by radiologists (Fig.~\ref{discussion_casestudy}b). For example, in a case where only Radiologist C reported retrocardiac opacity, GazeX identifies the same region by focusing on areas most relevant to the viewing pattern of Radiologist C. To evaluate the extent to which GazeX complements radiological findings, we compared its outputs with those of individual radiologists and the baseline model using majority voting of radiology descriptions as the reference standard (Fig.~\ref{discussion_casestudy}c). The results show that GazeX achieves higher precision and lower false negative rates compared with the baseline model, and its true positive count approaches that of three radiologists while surpassing that of Radiologist 4. These findings indicate that GazeX can recover additional clinically relevant observations that may be overlooked by individual experts.

\subsection{Enhancing diagnostic detail and spatial reasoning with inspection-guided modeling}
As shown in Fig.~\ref{discussion_casestudy}d, GazeX generates radiology reports that are more precise and grounded in image evidence compared to the baseline. By incorporating radiologists’ gaze areas paired with their descriptive reports, GazeX integrates fine-grained spatial information and mitigates common errors such as missing detail descriptions or hallucinated findings.

In case p15114531, the baseline Qwen2-VL mentions only anterior fixation hardware, omitting spatial context. GazeX, in contrast, produces a description consistent with the ground truth by recognizing surgical hardware in the lower cervical spine and confirming the absence of free air below the right hemidiaphragm. In case p1945978, Qwen2-VL merely states the presence of a PICC line, whereas GazeX provides a more complete description aligned with the ground truth, detailing its course and tip position. In case p14841168, Qwen2-VL hallucinates an increase in lung volume, contradicting radiological evidence of atelectasis. GazeX avoids this error and produces a report consistent with the ground truth, correctly identifying pleural effusion and atelectasis.

In contrast, the baseline model trained only on paired image–text data often emphasizes lexical similarity rather than spatial grounding, which leads to vague or incomplete descriptions. Even when constrained to designated regions of interest, the baseline may overlook findings that do not consistently manifest within those predefined areas. By integrating inspection-guided supervision, GazeX strengthens the connection between textual descriptions and actual visual evidence, enabling outputs that are more observation-driven and clinically reliable.

Fig.~\ref{discussion_casestudy}e and f further illustrate GazeX’s advantage in visual question answering and visual grounding tasks. In clinical question answering (Fig.~\ref{discussion_casestudy}e), GazeX consistently matches ground truth answers where other models fail, accurately capturing subtle distinctions in abnormality type, severity, and location. In phrase grounding evaluation (Fig.~\ref{discussion_casestudy}f), GazeX demonstrates precise bounding-box predictions that closely align with annotated ground truth, achieving high IoU scores. These results confirm that inspection pattern modeling not only improves narrative report generation, but also enhances spatial reasoning and targeted retrieval of image evidence.

Our results demonstrate that radiologists follow consistent and structured inspection strategies and that these patterns can be effectively captured and reproduced by GazeX. By integrating gaze-derived spatial and temporal information into its architecture, GazeX aligns its viewing behavior with expert readers and leverages this alignment to improve diagnostic accuracy, enrich spatial detail, and reduce clinically relevant errors. Importantly, this inspection-guided design allows model outputs to be examined in relation to familiar reading steps and anatomical regions, enabling clinicians to verify findings, assess coverage of critical areas, and identify potential omissions during routine review. In this way, GazeX complements individual radiologists by highlighting clinically relevant findings that may be overlooked while maintaining compatibility with established reporting workflows. Together, these results illustrate how inspection-guided modeling can bridge perceptual and interpretive processes, supporting AI systems that not only perform efficiently but also operate in a manner consistent with expert clinical reasoning.

\subsection{Clinical utility and pathway to uptake}

GazeX is designed for clinician-in-the-loop use rather than autonomous diagnosis or reporting. In current clinical practice, fully autonomous chest X-ray reporting remains uncommon, with most approved systems functioning as assistive or triage tools. Consistent with this observation, the inspection behavior learned by GazeX closely aligns with expert clinical practice and achieves performance comparable to inter‑radiologist variability. Quantitative analysis shows that the agreement between GazeX and radiologists in spatial and temporal inspection patterns falls within the range observed among experts themselves (Fig.~\ref{discussion_track} and Fig.~\ref{discussion_model}), supporting a deployment paradigm in which model outputs are reviewed and verified by clinicians rather than acted upon autonomously.

A central practical contribution of GazeX is its enforcement of systematic image inspection. Radiologists routinely follow structured reading protocols such as ABCDEF when interpreting chest X-rays, but they do not always explicitly report all normal or negative findings. GazeX models inspection trajectories directly, promoting comprehensive coverage of clinically relevant regions. It also produces structured visual evidence that can be audited step by step during clinical review. As shown in Fig.~\ref{discussion_track}a,c, radiologists exhibit consistent spatial coverage patterns across key thoracic regions. Fig.~\ref{discussion_model}a,b further shows that GazeX reproduces these spatial and temporal inspection strategies with fidelity comparable to expert-expert agreement. Together, these properties support coverage assurance, structured verification of model reasoning, and improved trust calibration, enabling clinicians to assess whether conclusions are grounded in appropriate visual evidence.

These inspection-guided behaviors translate into concrete practical outcomes. Case-level analysis indicates that GazeX reduces clinically relevant failure modes observed in baseline vision--language models, including hallucinated findings and omissions (Fig.~\ref{discussion_casestudy}c,d). Moreover, by producing finding-linked localized regions, GazeX enables actionable communication of results. Quantitative grounding improvements (Fig.~\ref{downstream_compare}c) and qualitative examples (Fig.~\ref{discussion_casestudy}f) demonstrate that the model can directly answer clinically meaningful questions such as lesion location, supporting clinician–AI communication during verification and clinician–clinician communication by making lesion location easy to reference.

In practical use, GazeX may serve as an assistive verification tool in situations where radiologists remain uncertain after an initial interpretation. By highlighting image regions associated with diagnostic ambiguity, GazeX, having learned from the visual inspection patterns of multiple experts and emulating how other radiologists would assess these regions in an X‑ray, provides a reference for second‑pass review. This enables clinicians to revisit specific anatomical areas during follow‑up inspection and cross‑checking, while preserving their clinical judgment. GazeX is intended as a decision‑support tool that proposes findings accompanied by linked visual evidence for clinician verification and editing and is not intended to generate final reports as a stand-alone diagnostic model.

\subsection{Limitations and Future Directions}

Building on these advances, future work will focus on evaluating the practical impact of inspection-guided evidence in clinical use. First, although GazeX is designed to support systematic image review, its effects on reporting efficiency, miss rates, and false-positive rates have not yet been quantified and will need to be assessed in prospective reader studies. Future studies will also examine trust calibration, including how clinician confidence aligns with case difficulty when inspection cues are provided, to better understand how such evidence supports human verification in practice, especially for the cases mentioned in Supplementary Fig.~{\color{blue} 17}. Second, our evaluation dataset does not include rare disease cases, which may limit robustness when encountering uncommon conditions in real-world practice. Expanding the dataset to cover a broader spectrum of pathologies represents an important direction for improving generalization. Third, the current framework does not incorporate external medical knowledge sources, such as retrieval-augmented generation (RAG), which could provide additional clinical context to guide inspection and interpretation. Finally, this study is restricted to chest X-rays, and the GazeX architecture has not yet been adapted or validated for other imaging modalities such as CT, MRI, or ultrasound. Future work will explore integrating domain knowledge in an auditable manner and extending the inspection-guided framework to additional modalities.

\begin{figure}[!t]
    \centering
    \includegraphics[width=1\columnwidth]{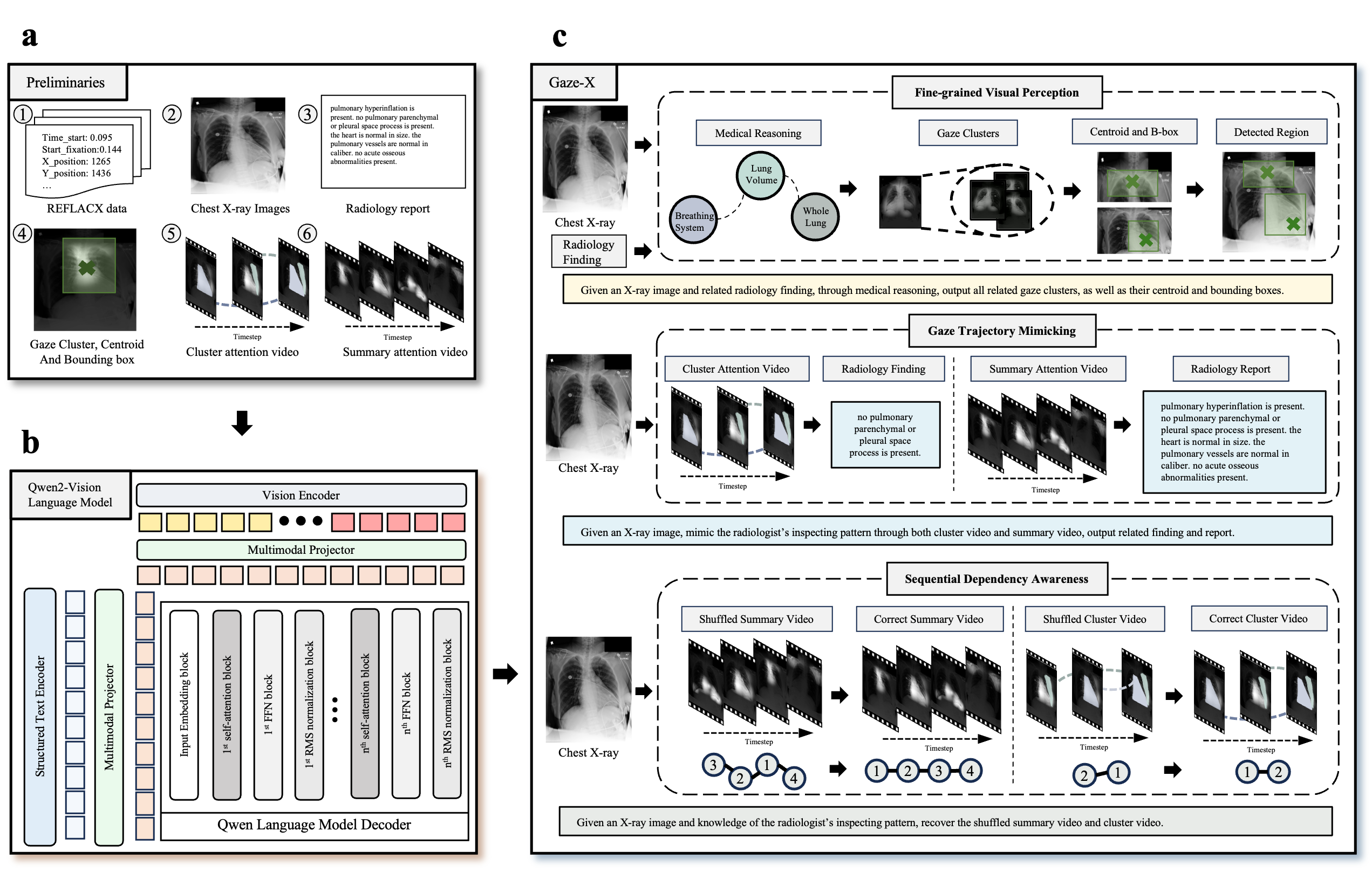}
    \caption{\textbf{Overview of the GazeX framework and its components.} \textbf{a}, Preliminaries: The REFLACX dataset provides synchronized radiology reports, frontal chest X-ray images, and eye-tracking data. Gaze clusters are extracted from fixation points, with each cluster characterized by its centroid and bounding box. These are assembled into cluster attention videos (per finding) and a summary attention video (aggregating all findings). \textbf{b}, GazeX architecture: Built on the Qwen2-Vision-Language backbone with optimized structured text encoder, vision encoder, and multimodal projector. \textbf{c}, Core modules: Fine-grained Visual Perception (FVP): Maps textual findings to gaze clusters, centroids, and bounding boxes via medical reasoning, enabling local-level spatial grounding. Gaze Trajectory Mimicking (GTM): Trains the model to follow radiologist-like inspection patterns using both cluster attention videos and summary attention videos. Sequential Dependency Awareness (SDA): Recovers the correct temporal order of shuffled attention videos, fostering global-level interpretive reasoning.}
    \label{method_fig}
\end{figure}

\section{Methods}
\subsection{Dataset for pre-training} 
The REFLACX dataset contains 3,032 synchronized eye-tracking and report transcription cases, for a total of 2,616 unique frontal chest X-ray images from the MIMIC-CXR dataset, each read by one or more of five board-certified radiologists. During each case, radiologists' descriptive findings and eye movements were recorded while they viewed and dictated X-ray interpretations, with eye tracking captured continuously at 1,000 Hz. Data were collected in three phases. Phases 1 and 2 involve multiple radiologists viewing the same X-rays. Phase 1 used 59 chest X-rays, and Phase 2 used 50 chest X-rays. Each X-ray was interpreted by the same group of radiologists (up to 5). This produced 285 and 240 eye-tracking recordings for Phases 1 and 2, respectively, with some recordings being dropped due to technical issues. Phase 3 represents the main component of the dataset, including 2,507 unique cases, and each radiologist reviewed a different set of images. For training and evaluation, we adopt an 80/20 split over the full set of 3,032 recordings. Specifically, we use all 240 cases from Phase 2 and randomly sample 350 from Phase 3 to form a 590-case test set ($\sim$20\%), while the remaining 2,442 cases are used for training.

\begin{figure}[!t]
    \centering
    \includegraphics[width=1\columnwidth]{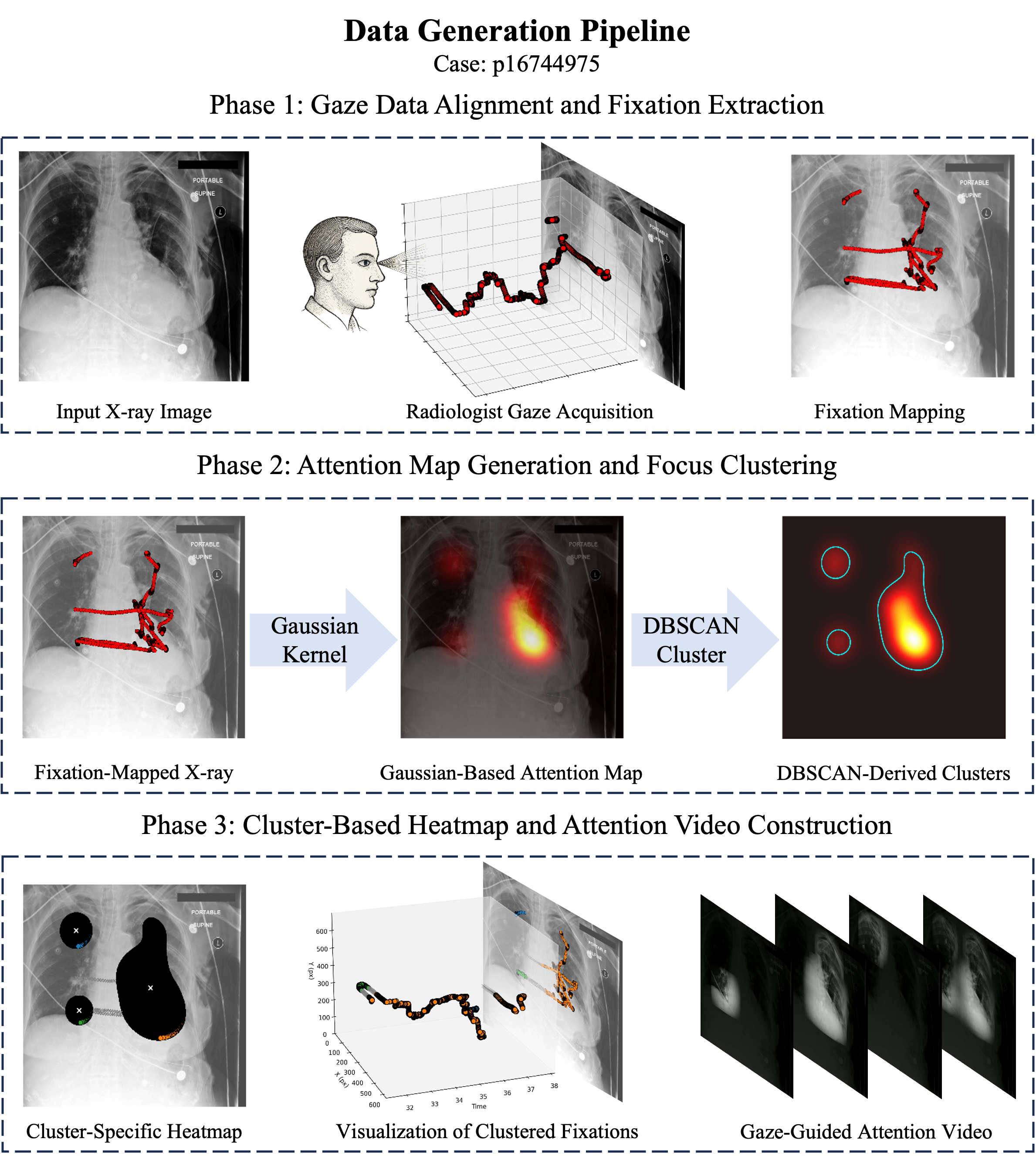}
    \caption{\textbf{Illustrative example of the data processing pipeline.} Given a chest X-ray image, following the pipeline defined by the pseudocode Algorithm~{\color{blue}1} in supplementary data, gaze-guided attention videos are generated through three phases for model training. Phase 1: Acquire radiologist gaze records and extract fixations based on their verbal descriptions of the regions of interest. Phase 2: For each specific verbal description, apply Gaussian attention map to the corresponding fixation data, followed by DBSCAN clustering to identify attention clusters. Phase 3: Refine the extracted attention clusters and construct the final gaze-guided attention videos for use in model training.}
    \label{illustrative_example}
\end{figure}

\paragraph{Data preparation.}
All chest X-ray images in REFLACX are downsampled by a factor of 4 along both spatial dimensions (that is, width and height), resulting in a total area scaling of 1/16. The gaze coordinates are correspondingly scaled by a factor of 4 to remain aligned with the resized images. For each verbal mention of an abnormality, we extract the set of fixation points $F_i=(x_i, y_i, t_i)$ occurring within a 2.5-second window prior to the timestamp of the mention. Each fixation point is used to generate a 2D Gaussian heatmap $G$ centered at $(x_i, y_i)$ with standard deviations $\sigma_x$ and $\sigma_y$ along the horizontal and vertical axes, respectively. The values of $\sigma_x$ and $\sigma_y$ are derived from the angular resolution of the image at the fixation point, expressed in pixels per degree of visual angle (that is, angular resolution x pixels per degree and angular resolution y pixels per degree). Each Gaussian is evaluated over a bounded window with radius 3 times of $\text{max}(\sigma_x, \sigma_y)$, which captures more than 99\% of the Gaussian mass. The resulting Gaussians are summed to produce a temporally aligned spatial attention map for each finding. We then apply DBSCAN clustering \cite{DBscan} ($\epsilon=0.3$, MinPts $=10$) to the resulting attention maps to isolate distinct regions of visual focus. Here, $\epsilon$ defines the maximum distance between points to be considered neighbors, and MinPts is the minimum number of points required to form a dense region, allowing the algorithm to identify focused areas while filtering out sparse noise. For each cluster, we apply the previously described method to the corresponding coordinates in order to generate a new heatmap. The centroid of each cluster is subsequently computed via intensity-weighted averaging of the corresponding coordinates, while bounding box anchors are derived from the maximum and minimum XY coordinates of the top 20\% highest-intensity pixels within the heatmap. These clusters are used to construct an attention video in which the clusters are sequentially overlapped with the resized chest X-ray. Each finding with $K$ clusters contributes $K+1$ frames to their attention video $V_m$, one for each cluster in temporal order and one summary frame $H_{total}$ showing all clusters for that finding. Suppose a report contains $M$ findings, with each finding $m$ associated with video $V_m$ containing $K+1$ frames. Then the total number of frames in the summary attention video $V$ is given by $\sum_{i=1}^{|M|} (K_i + 1)$. An illustration of the REFLACX data, chest X-ray image, radiology report, gaze group, cluster's centroid $c$ and bounding box $b$, cluster video $V_m$ of a description and summary video $V$ of the entire diagnosis on a chest X-ray image is provided in Fig.~\ref{method_fig}a and the pipeline for generating attention videos is illustrated in Fig.~\ref{illustrative_example}.

\subsection{GazeX}
We propose GazeX (Fig.~\ref{method_fig}), a 2 billion-parameter foundation model for chest X-rays that builds on the Qwen2-Vision-Language architecture, with optimized structured text encoder, vision encoder, and multimodal projector components. To enable spatially grounded and context-aware understanding, GazeX integrates three modules to capture different levels of visual and semantic complexity: fine-grained visual perception (FVP), gaze trajectory mimicking (GTM), and sequential dependency awareness (SDA).

In FVP, we construct a medical reasoning dataset that pairs each radiology finding with anatomical locations, which are further refined into gaze-based clusters. The model is trained to predict these clusters along with their centroids and bounding boxes, allowing it to establish precise spatial correspondences between textual findings and visual evidence in the X-ray. GTM enables GazeX to learn radiologist-like inspection patterns by supervising it with cluster attention videos $v_i$ aligned with specific findings and summary attention videos $V$ aligned with complete radiology reports. This guides the model to attend to both localized and holistic regions of the chest X-ray in a manner consistent with clinical reading behavior. Finally, in SDA, we shuffle the temporal order of both $v_i$ and $V$ and train GazeX to recover the correct viewing sequence, promoting sensitivity to the temporal structure of radiological workflows and enhancing global contextual understanding. Details of the data processing pipeline are provided in Supplementary Algorithm~{\color{blue} 1}.

\subsection{Downstream evaluation tasks}
To assess the effectiveness of GazeX across different levels of visual-linguistic understanding, we evaluate its performance on three downstream tasks of decreasing complexity: radiology report generation, visual question answering and visual grounding. These tasks are designed to investigate complementary capabilities of the model, ranging from language generation to contextual interpretation and spatial localization. Report generation evaluates global-level understanding by requiring the model to synthesize coherent radiological summaries that integrate information throughout the entire X-ray image. Visual question answering assesses local-level comprehension by testing the model’s ability to answer fine-grained clinical questions about visual content. Visual grounding targets image-level understanding, which requires aligning free-text phrases with specific anatomical regions. 

\subsection{Datasets and Comparsion Methods for downstream evaluation tasks}
For all datasets, we adhere to their official training, validation, and test splits. All chest X-ray images and the corresponding bounding box coordinates are downsampled by a factor of 4 along both spatial dimensions (that is, width and height), corresponding to a 1/16 reduction in spatial area.

\paragraph{Radiology report generation.}
We use two widely adopted datasets: MIMIC-CXR~\citep{mimicxr} and IU X-Ray~\citep{IUXray}. MIMIC-CXR contains 377,110 chest X-ray images paired with 227,835 free-text radiology reports, while IU X-Ray includes 7,470 images and 3,955 reports. For both cases, we target the "Findings" section of each report for generation. We selected representative chest X-ray report generation models for this evaluation, including R2Gen~\citep{r2gen}, CMN~\citep{chen-etal-2021-cross-modal}, MA~\citep{tmiauto}, RGRG~\citep{rgrg}, R2GenGPT~\citep{WANG2023100033}, PromptMRG~\citep{promptmrg}, STREAM~\citep{Stream}, and our proposed GazeX.

\paragraph{Visual question answering.}
We use the Medical-CXR-VQA dataset~\citep{medical_vqa_dataset}, which includes 780,014 question–answer pairs based on MIMIC-CXR images. The questions span six categories: abnormality (190,525 pairs), location (104,680 pairs), type (69,486 pairs), severity level (111,715 pairs), view (92,048 pairs) and presence (211,560 pairs), supporting both visual and linguistic reasoning. We selected both models fine-tuned on the Medical-CXR-VQA dataset and zero-shot foundation models, including LLaVA-Med-7B~\citep{li2023llava}, RadFM-14B~\citep{RadFM}, HuatuoGPT-Vision-7B~\citep{chen2024huatuogpt}, HuatuoGPT-Vision-32B~\citep{chen2024huatuogpt}, Bio-Medical-MultiModal-Llama-8B~\citep{ContactDoctor_MEDLLM}, Medical-CXR-VQA~\citep{medical_vqa_dataset}, MedGemma-4B-it~\citep{medgemma-hf}, Lingshu-7B~\citep{xu2025lingshu}, and Lingshu-32B~\citep{xu2025lingshu}. We tested model variance across different parameter sizes (in billions) to compare the trade-off between model parameter size and performance, since smaller models can be more feasible for real-world implementation with limited resources.

\paragraph{Visual grounding.} 
We use the MS-CXR dataset~\citep{ms-cxr}, which provides 1,153 image–sentence pairs consisting of bounding boxes and associated phrases derived from MIMIC-CXR. These structured annotations enable precise grounding of visual regions to fine-grained textual descriptions. We selected models with different training methods, including ChEX~\citep{chex} (visual grounding foundation model), TransVG~\citep{transvg} (supervised learning model for visual grounding), BioVIL~\citep{ms-cxr} (CLIP-based learning model for visual grounding), and OmniFM-DR~\citep{Omnifm-dr} (generative visual-grounding model).

\subsection{Evaluation metrics}
\paragraph{Radiology report generation.}
Performance is assessed using standard natural language generation metrics, including BLEU-1 to BLEU-4~\citep{papineni2002bleu}, METEOR~\citep{lavie-agarwal-2007-meteor}, and ROUGE-L~\citep{lin2004rouge}, to measure textual similarity. In addition, we use the CheXbert~\citep{chexbert} labeler to assign reports to 14 common disease categories to provide an assessment of the clinical accuracy of the generated text. For model comparison with existing approaches, we adopt the mainstream evaluation methodology that employs F1 scores, where only positive findings are classified as positive cases. In the ablation study, we observe that the pretraining dataset contains radiologist-generated findings characterized by inherent uncertainty, which our model subsequently learns to replicate, thereby reflecting the diagnostic uncertainty present in radiological examinations. To comprehensively evaluate this behavior, we implement a multiclass classification framework.

\paragraph{Visual question answering.} 
We report top1 phrase-matching accuracy~\citep{medical_vqa_dataset}. Predictions are decomposed into phrases and are considered correct if any predicted phrase exactly matches any ground-truth phrase.

\paragraph{Visual grounding.}  
We evaluate this task using mean Intersection over Union (mIoU) and mean Average Precision (mAP)~\citep{ms-cxr}. For mIoU, predicted bounding boxes are selected based on confidence scores, merged into a binary mask, and compared to a merged ground truth mask using IoU thresholds of \{0.1, 0.2, 0.3, 0.4, 0.5\}. IoU is first averaged across all sentences for each threshold and then averaged across thresholds. For mAP, the task is treated as object detection, with each sentence corresponding to a unique class. The average precision is computed using IoU thresholds of \{0.1, 0.2, 0.3, 0.4, 0.5, 0.6, 0.7\} and averaged across all classes and thresholds.

\newpage
\bibliographystyle{myrecomb}
\bibliography{mybib}

\section{Data availability}
All the datasets we use are listed in Methods and are publicly available. MIMIC-CXR and IU X-ray are publicly available at 
\url{https://physionet.org/content/mimic-cxr/2.1.0/} and  
\url{https://www.kaggle.com/datasets/raddar/chest-xrays-indiana-university} respectively. 
The Medical-CXR-VQA dataset is publicly accessible at 
\url{https://physionet.org/content/medical-cxr-vqa-dataset/1.0.0/}. 
The MS-CXR images and annotations are publicly accessible at 
\url{https://physionet.org/content/ms-cxr/1.1.0/}. 
The dataset REFLACX for reports and eye-tracking data in chest x-rays is accessible at 
\url{https://physionet.org/content/reflacx-xray-localization/1.0.0/}.

\section{Code availability}
The code for preprocessing and reproducing the results is in the GitHub repository: \href{https://github.com/KennyLee123/GazeX.git}{GazeX}.

\section{Acknowledgements} 
Guang Yang was supported in part by the ERC IMI (101005122), the H2020 (952172), the MRC (MC/PC/21013), the Royal Society (IEC/NSFC/211235), the NVIDIA Academic Hardware Grant Program, the SABER project supported by Boehringer Ingelheim Ltd, NIHR Imperial Biomedical Research Centre (RDA01), The Wellcome Leap Dynamic resilience program (co-funded by Temasek Trust)., UKRI guarantee funding for Horizon Europe MSCA Postdoctoral Fellowships (EP/Z002206/1), UKRI MRC Research Grant, TFS Research Grants (MR/U506710/1), Swiss National Science Foundation (Grant No. 220785), and the UKRI Future Leaders Fellowship (MR/V023799/1, UKRI2738). Peiyuan Jing is supported by the Swiss National Science Foundation (SNSF) under grant number 20HW-1 220785. Zhenxuan Zhang was supported by a CSC Scholarship.

\section{Author contributions}

K.L. conceived the research.
K.L., P.J., Y.Y. and Z.Z. implemented the models.
K.L., P.J., Y.Y. and Z.Z. ran the experiments and interpreted the results.
K.L., P.J., Y.F., T.W. and Z.Z. designed the downstream experiments.
K.L., P.J. and Z.Z. organized and processed the publicly available data.
K.L., P.J. and Z.Z. visualized the experimental results.
G.Y. supervised the research.
All authors wrote, read and approved the final manuscript.


\section{Competing interests}
The authors declare no competing interests.




\end{document}